\documentclass[sigconf]{acmart}

\usepackage{booktabs} 
\usepackage{xspace}
\usepackage{float}
\usepackage{natbib}
\usepackage{algorithm}
\usepackage{caption}
\usepackage[noend]{algpseudocode}
\usepackage{comment}
\usepackage{subfigure}
\usepackage{etoolbox}
\usepackage{placeins}
\usepackage{dblfloatfix}
\makeatletter
\renewcommand{\thesubfigure}{\alph{subfigure}}
\renewcommand{\@thesubfigure}{(\thesubfigure)\hskip\subfiglabelskip}
\makeatother

\newcommand{\meesexplore}{\textsc{ME-ES explore}\xspace}
\newcommand{\meesexploit}{\textsc{ME-ES exploit}\xspace}
\newcommand{\mees}{\textsc{ME-ES}\xspace}
\newcommand{\nses}{\textsc{NS-ES}\xspace}
\newcommand{\nsraes}{\textsc{NSRA-ES}\xspace}
\newcommand{\nsres}{\textsc{NSR-ES}\xspace}
\newcommand{\es}{\textsc{ES}\xspace}
\newcommand{\nes}{\textsc{NES}\xspace}
\newcommand{\ga}{\textsc{GA}\xspace}
\newcommand{\gas}{\textsc{GA}s\xspace}
\newcommand{\mboa}{\textsc{M-BOA}\xspace}
\newcommand{\imgep}{\textsc{IMGEP}\xspace}
\newcommand{\nslc}{\textsc{NSLC}\xspace}
\newcommand{\ns}{\textsc{NS}\xspace}

\newcommand{\cmaes}{\textsc{CMA-ES}\xspace}
\newcommand{\cmame}{\textsc{CMA-ME}\xspace}

\newcommand{\mega}{\textsc{ME-GA}\xspace}
\newcommand{\me}{\textsc{MAP-E}lites\xspace}
\newcommand{\qd}{\textsc{QD}\xspace}
\newcommand{\meesexex}{\textsc{ME-ES explore-exploit}\xspace}
\definecolor{myred}{rgb}{0.8,0,0}
\definecolor{deepForestGreen}{rgb}{.2,.8,.04}
\definecolor{magenta}{rgb}{0.7,0,1}

\copyrightyear{2020}
\acmYear{2020}
\setcopyright{acmcopyright}\acmConference[GECCO '20]{Genetic and Evolutionary Computation Conference}{July 8--12, 2020}{Cancún, Mexico}
\acmBooktitle{Genetic and Evolutionary Computation Conference (GECCO '20), July 8--12, 2020, Cancún, Mexico}
\acmPrice{15.00}
\acmDOI{10.1145/3377930.3390217}
\acmISBN{978-1-4503-7128-5/20/07}

\begin{document}
\title{Scaling MAP-Elites to Deep Neuroevolution}

\author{C\'edric Colas}
\affiliation{%
  \institution{INRIA}
}
\authornote{Corresponding author: cedric.colas@inria.fr. Work conducted during an internship at Uber AI Labs. All authors were affiliated to Uber AI Labs at the time.}

\author{Joost Huizinga}
\affiliation{%
  \institution{Uber AI Labs}
}
\author{Vashisht Madhavan}
\affiliation{%
  \institution{Element Inc.}
}

\author{Jeff Clune}
\affiliation{%
  \institution{OpenAI}
}


\begin{abstract}
Quality-Diversity (QD) algorithms, and MAP-Elites (ME) in particular, have proven very useful for a broad range of applications including enabling real robots to recover quickly from joint damage, solving strongly deceptive maze tasks or evolving robot morphologies to discover new gaits. However, present implementations of ME and other QD algorithms seem to be limited to low-dimensional controllers with far fewer parameters than modern deep neural network models. In this paper, we propose to leverage the efficiency of Evolution Strategies (ES) to scale MAP-Elites to high-dimensional controllers parameterized by large neural networks. We design and evaluate a new hybrid algorithm called MAP-Elites with Evolution Strategies (ME-ES) for post-damage recovery in a difficult high-dimensional control task where traditional ME fails. Additionally, we show that ME-ES performs efficient exploration, on par with state-of-the-art exploration algorithms in high-dimensional control tasks with strongly \emph{deceptive} rewards.
\end{abstract}

%
%



\maketitle

\section{Introduction}

The path to success is not always a straight line. This popular saying describes, in simple terms, a key problem of non-convex optimization. When optimizing a model for a non-convex objective function, an algorithm that greedily follows the gradient of the objective might get stuck in local optima. One can think of an agent in a maze with numerous walls. An algorithm that minimizes the distance between the position of the agent and the maze center will surely lead to the agent getting stuck in a corner. Based on this observation, \citet{ns} proposed a thought-provoking idea: ignore the objective and optimize for novelty instead. Novelty search (\ns) continually generates new behaviors without considering any objective and, as such, is not subject to the local optima encountered by algorithms following fixed objectives. In cases like our maze example, \ns can lead to better solutions than objective-driven optimization \cite{ns}.

Despite the successes of \ns, objectives still convey useful information for solving tasks. As the space of possible behaviors increases in size, \ns can endlessly generate novel outcomes, few of which may be relevant to the task at hand. Quality-Diversity (\qd) algorithms address this issue by searching for a collection of solutions that is both diverse and high-performing~\cite{nslc, mapelite, cully_qd}. \me, in particular, was used to generate diverse behavioral repertoires of walking gaits on simulated and physical robots, which enabled them to recover quickly from joint damage \cite{cully2015robots}. 

In \qd algorithms like \me or \nslc, a Genetic Algorithm (\ga) is often used as the underlying optimization algorithm. A candidate controller is selected to be mutated and the resulting controller is evaluated in the environment, leading to a performance measure (fitness) and a behavioral characterization (low-dimensional representation of the agent's behavior). \qd algorithms usually curate some form of a \textit{behavioral repertoire}, a collection of high-performing and/or diverse controllers experienced in the past (sometimes called \textit{archive} \cite{nslc, cully_qd} or \textit{behavioral map} \cite{mapelite}). Each newly generated controller can thus be added to the behavioral repertoire if it meets some algorithm-dependent conditions. From a parallel line of work, Intrinsically Motivated Goal Exploration Processes (\imgep) also curate behavioral repertoires and are often based on \textsc{GA}-like optimizations \cite{baranes2013active, imgep}. Agents are able to set their own goals in the behavioral space, and try to reach them by combining controllers that reached behavioral characterizations ($BC$s) close to these goals. Uniform goal selection here triggers a novelty effect where controllers reaching sparse areas are used more often, while learning-progress-based sampling implements a form of \qd where quality is defined as the ability to reliably achieve goals \cite{baranes2013active}. 

Thus far, the most successful demonstrations of \qd algorithms have been on robotics problems with relatively simple, low-di\-men\-sio\-nal controllers \cite{ns,nslc,mapelite,cully_qd}. Modern robot controllers such as ones based on neural networks can have millions of parameters, which makes them difficult --though not impossible-- to optimize with Evolutionary Algorithms \cite{deepga}. Such large controllers are usually trained through Deep Reinforcement Learning (DRL), where an agent learns to perform a sequence of actions in an environment so as to maximizes some notion of cumulative reward \cite{sutton2018reinforcement}. DRL is concerned with training deep neural networks (DNNs), typically via stochastic gradient descent (SGD) to facilitate learning in robot control problems. Unlike in supervised learning, training data in DRL is generated by having the agent interact with the environment. If the agent greedily takes actions to maximize reward --a phenomenon known as \textit{exploitation}-- it may run into local optima and fail to discover alternate strategies with larger payoffs. To avoid this, RL algorithms also need \textit{exploration}. RL algorithms usually explore in the action space, adding random noise on a controller's selected actions ($\epsilon$-greedy) \cite{mnih2013playing,lillicrap2015continuous,fujimoto2018addressing}. More directed exploration techniques endow the agent with \textit{intrinsic motivations}  \cite{schmidhuber1991possibility, oudeyer2007intrinsic, barto2013intrinsic}. Most of the time, the reward is augmented with an \textit{exploration bonus}, driving the agent to optimize for proxies of \textit{uncertainty} such as novelty \cite{countbased}, prediction error \cite{pathak2017curiosity, burda2018exploration}, model-disagreement \cite{shyam2018model}, surprise \cite{achiam2017surprise} or expected information gain \cite{houthooft2016vime}.


In contrast with \qd algorithms however, DRL algorithms are usually concerned with training a single controller to solve tasks \cite{mnih2013playing,lillicrap2015continuous,fujimoto2018addressing,pathak2017curiosity}. The agent thus needs to rely on a single controller to adapt to new environments \cite{nichol2018gotta,portelas2019teacher}, tasks \cite{finn2017model}, or adversarial attacks \cite{gleave2019adversarial}. \citet{kume2017map} outlines perhaps the first RL algorithm to form a behavioral repertoire by training multiple policies. Optimizing for performance, this algorithm relies on the instabilities of the underlying learning algorithm (Deep Deterministic Policy Gradient \cite{lillicrap2015continuous}) to generate a diversity of behaviors that will be collected in a \textit{behavioral map}. Although the performance optimization of this \qd algorithm leverages DRL, exploration remains incidental.

In recent years, Deep Neuroevolution has emerged as a powerful competitor to SGD for training DNNs. \citet{es}, in particular, presents a scalable version of the Evolution Strategies (\es) algorithm, achieving performances comparable to state-of-the-art DRL algorithms on high-dimensional control tasks like Mujoco \cite{brockman2016openai} and the Atari suite \cite{bellemare2013arcade}. Similarly, Genetic Algorithms (\ga) were also shown to be capable of training DNN controllers for the Atari suite, but failed to do so on the Mujoco suite \cite{deepga}. \es, in contrast with \ga, combines information from many perturbed versions of the parent controller to generate a new one. Doing so allows the computation of gradient estimates, leading to efficient optimization in high-dimensional parameter spaces like DNNs \cite{es}. Recent implementations also utilize the resources of computing clusters by parallelizing the controller evaluations, making \es algorithms competitive with DRL in terms of training time \cite{es}. Since \citet{deepga} showed that even powerful, modern \gas using considerable amounts of computation could not solve continuous control tasks like those from the Mujoco suite, we propose to unlock \qd for high-dimensional control tasks via a novel QD-\es hybrid algorithm. Doing so, we aim to benefit from the exploration abilities and resulting behavioral repertoires of \qd algorithms, while leveraging the ability of \es to optimize large models.

\nses made a first step in this direction by combining \ns with \es: replacing the performance objective of \es with a novelty objective~\cite{nses}. Two variants were proposed to incorporate the performance objective: \nsres, which mixes performance and novelty objectives evenly and \nsraes, which adaptively tunes the ratio between the two. These methods --like most RL exploration methods-- use a mixture of exploitation and exploration objectives as a way to deal with the \textit{exploitation-exploration tradeoff}. While this might work when the two objectives are somewhat aligned, it may be inefficient when they are not \cite{pugh2016searching}. Several works have started to investigate this question and some propose to disentangle exploration and exploitation into distinct phases \cite{geppg,beyer2019mulex,zhang2019scheduled}. \qd presents a natural way of \emph{decoupling} the optimization of exploitation (quality) and exploration (diversity) by looking for high-performing solutions in local niches of the behavioral space, leading to local competition between solutions instead of a global competition \cite{nslc,mapelite,cully_qd}.

\paragraph{Contributions} 
In this work, we present \mees, a version of the powerful \qd algorithm \me that scales to hard, high-dimensional control tasks by leveraging \es. Unlike \nses and its variants, which optimize a small population of DNN controllers (e.g.~$5$) \cite{nses}, \mees builds a large repertoire of diverse controllers. Having a set of diverse, high-performing behaviors not only enables efficient exploration but also robustness to perturbations (either in the environment itself or in the agent's abilities). In behavioral repertoires, the burden of being robust to perturbations can be shared between different specialized controllers, one of which can solve the new problem at hand. Algorithms that train a single controller, however, need this controller to be robust to all potential perturbations. 
We present two applications of \mees. The first is damage recovery in a high-dimensional control task: after building a repertoire of behaviors, the agent is damaged (e.g. disabled joints) and must adapt to succeed. We show that \mees can discover adaptive behaviors that perform well despite the damage, while a previous implementation of \me based on \ga fails. The second application is exploration in environments with strongly deceptive rewards. We show that agents trained with \mees perform on par with state-of-the-art exploration algorithms (\nses and variants). Table~\ref{tab:related} presents a classification of \mees and related algorithms presented in this paper along three dimensions: 1) whether they use pure exploration, pure exploitation or both, 2) whether they rely on gradient-based or mutation-based learning algorithms and 3) whether they couple or decouple the trade-off between exploration and exploitation (if applicable). As \es computes \textit{natural gradient} estimations through Monte-Carlo approximations \cite{nes}, we refer to \es-powered methods as \textit{gradient-based}. 

\begin{table}
  \caption{\textbf{Classification of related algorithms}}
  \label{tab:related}
  \begin{tabular}{c|cc}
        \multicolumn{1}{c|}{} & Gradient-Based &  Mutation-Based\\
                        
    \midrule
     Pure               & \small \es                    & \small \ga \\
     Exploitation       & & \\ \hline
     Pure               & \small \nses                  & \small \ns \\ 
     Exploration        &                        & \\ \hline 
     Exploration \&        & \small Coupled: \nsres, \nsraes      & \small \mega \\
     Exploitation       & \small \textbf{Decoupled: \mees} & \\
\end{tabular}
\end{table}


\section{Background}
\subsection{MAP-Elites}
\me was first presented in \citet{mapelite}. In addition to an objective function --or \emph{fitness} function $(F)$-- which measures the performance of an agent, \me assumes the definition of a behavioral characterization $(BC)$ mapping the state-action trajectory of an agent in its environment to a low-dimensional embedding lying in a \textit{behavioral space}. \me keeps track of an archive of individuals (controllers) tried in the past along with their associated fitness values and behavioral characterizations. The aim is to curate a repertoire of behaviorally diverse and high-performing agents. To this end, the behavioral space is discretized into \textit{cells} representing behavioral \textit{niches}, where each niche maintains the highest performing individual whose behavioral characterization falls into its cell. Individuals in each niche optimize for the fitness objective, yet implicitly have a pressure for diversity, as they are driven towards empty and under-optimized cells to avoid the selection pressure of cells with highly optimized individuals.

After initializing the archive with a few randomly initialized controllers, \me repeats the following steps:
\begin{enumerate}
    \item Select a populated cell at random,
    \item Mutate the cell's controller (\ga) to obtain a new controller,
    \item Gather a trajectory of agent-environment interactions with the new controller and obtain its fitness and $BC$,
    \item Update the archive: add the controller to the cell where the $BC$ falls if \textit{Rule 1}) the cell is empty or \textit{Rule 2}) the fitness is higher than that of the controller currently in the cell.
\end{enumerate}

The two rules guiding additions to the archive implement a decoupling between exploitation (quality) and exploration (diversity). \textit{Rule 1} implements exploration, as it ensures the preservation of controllers that exhibit novel behavior (i.e. mapped to empty cells). \textit{Rule 2} implements exploitation, as it enforces local competition and retains only the highest performing solution in each behavioral niche. In this manner, the exploration and exploitation objectives cannot contradict each other. We call the traditional implementation of \me based on \ga \textit{\mega} while \textit{\me} encompasses both \mega and \mees variants.

\subsection{Evolution Strategies}
Evolution Strategies (\es) is a class of black box optimization algorithms inspired by natural evolution \cite{renchengberg_es, essurvey}. For each generation, an initial parameter vector (the \textit{parent}), is mutated to generate a population of parameter vectors (the \textit{offspring}). The \textit{fitness} of each resultant offspring is evaluated and the parameters are combined such that individuals with high fitness have higher influence than others. As this process is repeated, the population tends towards regions of the parameter space with higher fitness, until a convergence point is reached.

We use a version of \es introduced in \citet{es}, which itself belongs to the subcategory of Natural Evolution Strategies (\nes) \cite{nes, sehnke2010parameter}. In \nes, the population of parameter vectors $\theta$ is represented by a distribution $p_\psi(\theta)$, parameterized by $\psi$. Given an objective function $F$, \nes algorithms optimize the expected objective value $\mathbb{E}_{\theta \sim p_\psi}F(\theta)$ using stochastic gradient ascent. Recently, \citet{es} proposed a version of the \nes algorithm able to scale to the optimization of high-dimensional parameters ($\approx 10^5$). In their work, they address the RL problem, for which $\theta$ refers to the parameters of a controller while the fitness function is the reward obtained by the corresponding controller over an episode of environment interactions. Given the parent controller parameters $\theta$, the offspring population $(\theta_{i} \: \forall i \in [1..n]$) is sampled from the isotropic multivariate Gaussian distribution $\theta_i\sim \mathcal{N}(\theta, \sigma^2I)$ with fixed variance $\sigma^2$. As in \textsc{reinforce\xspace} \cite{reinforce}, $\theta$ is updated according to the following gradient approximation:

$$\nabla_\psi \mathbb{E}_{\theta\sim p_\psi}[F(\theta)]\approx\frac{1}{n}\sum^n_{i=1}F(\theta_i)\nabla_\psi logp_\psi(\theta),$$
\noindent
where $n$ is the offspring population size, usually large to compensate for the high variance of this estimate. In practice, any $\theta_i$ can be decomposed as $\theta_i=\theta+\sigma\epsilon_i$, and the gradient is estimated by:

$$\nabla_\psi \mathbb{E}_{\epsilon\sim\mathcal{N}(0,I)}[F(\theta+\sigma\epsilon)]\approx\frac{1}{n\sigma}\sum^n_{i=1}F(\theta_i)\epsilon_i.$$

As in \citet{es}, we use virtual batch normalization of the controller's inputs and rank-normalize the fitness $F(\theta^i)$. The version of \nes used in this paper is strictly equivalent to the one proposed in \citet{es}. We simply refer to it as \es hereafter.


\section{Methods}
\subsection{ME-ES}
The \mees algorithm reuses the founding principles of \me leading to damage robustness and efficient exploration, while leveraging the optimization performance of \es. \mees curates a \textit{behavioral map} $(BM)$, an archive of neural network controllers parameterized by $\theta$. Every \textit{n\_optim\_gens} generations, a populated cell and its associated controller $\theta_{cell}$ are sampled from the archive. This controller is then copied to $\theta_g$, which is subsequently evolved for \textit{n\_optim\_gens} generations using \es. At each generation, offspring parameters $\theta_g^i \sim \mathcal{N}(\theta_g, \sigma^2I)$ are only used to compute an update for $\theta_{g}$. We only consider $\theta_g$ for addition to the $BM$ to ensure robust evaluations ($30$ episodes) and a fair comparison with competing algorithms such as \nses. Algorithm~\ref{alg:mees} provides a detailed outline of the way \mees combines principles from \me and \es.

\begin{algorithm}[!htb]
\caption{\mees Algorithm}\label{alg:mees}
\begin{algorithmic}[1]
\State \textbf{Input:} n\_gens, pop\_size, $\sigma$, n\_optim\_gens, empty behavioral map $BM$, n\_eval
\State \textbf{Initialize:} $BM \gets (\theta_0 \sim init_{normc}(), F(\theta_0), BC(\theta_0))$
\For{$g = [0, \:$n\_gens$]$}
    \If{g \% n\_optim\_gens == 0}
        \State mode $\gets$ explore\_or\_exploit()
        \If{mode == `explore'}
            \State $\theta_{cell} \gets$ sample\_explore\_cell()
        \ElsIf{mode == `exploit'}
            \State $\theta_{cell} \gets$ sample\_exploit\_cell()
        \EndIf
        \State $\theta_g \gets \theta_{cell}$
    \EndIf
    \State $\theta_g \gets$ ES\_optim$(\theta_g$, pop\_size, $\sigma$, objective=mode$)$
    \State $F(\theta_g)$, $BC(\theta_g)$ $\gets$ Evaluate($\theta_g$, n\_eval)
    \State Update\_BM$(\theta_g, \:F(\theta_g), BC(\theta_g))$
\EndFor
\end{algorithmic}
\end{algorithm}

\paragraph{\mees variants}We define three variants of \mees that differ in the objective optimized by \es:
\begin{itemize}
    \item  In \meesexploit, the objective is the fitness function $F$. $F(\theta)$ is computed by running the agent's controller in the environment for one episode (\textit{directed exploitation}). 
    \item In \meesexplore, the objective is the \textit{novelty} function $N(\theta, k)$, which we define as the average Euclidean distance between a controller's $BC$ and its $k$ nearest neighbors in an \emph{archive} storing all previous $\theta_g$, regardless of their addition to the $BM$ (one per generation). Because the novelty objective explicitly incentivizes agents to explore their environment, we call it a \textit{directed exploration} objective.
    \item Finally, \meesexex alternates between both objectives, thus implementing a decoupled exploration and exploitation procedure. Because it explicitly optimizes for both objectives, \meesexex conducts both \textit{directed exploration} and \textit{directed exploitation}.
\end{itemize}
Note that \meesexplore performs undirected exploitation; while the \es steps do not directly optimize for fitness, performance measures are still used to update the $BM$ (Rule 2 of the map updates). In the same way, \meesexploit also performs undirected exploration, as the $BM$ is updated with novel controllers (Rule 1 of map updates). Like \mega, all versions of \mees thus perform forms of exploration and exploitation. Only optimization steps with \es, however, enable agents to perform the \textit{directed} version of both exploitation and exploration.

\paragraph{Cell sampling}The number of generations performed by \mees is typically orders of magnitude lower than \me, as each generation of \mees involves a greater number of episodes of environment interaction ($10^4$ instead of $1$ for \mega). Thus, the sampling of the cell from which to initiate the next \textit{n\_optim\_gens} generations is crucial (see Algorithm~\ref{alg:mees}). Here, we move away from the cell selection via uniform sampling used in \me, towards biased cell sampling. We propose two distinct strategies. For exploitation steps, we select from cells with high fitness controllers, under the assumption that high fitness cells may lead to even higher fitness cells. For exploration steps, we select from cells with high novelty scores. By definition, novel controllers are in under-explored areas of the search space and mutating these controllers should lead to novel areas of the behavior space \cite{ns,nslc}. In practice, as the number of cells populated increases with each generation, the bias towards selecting cells with higher performance or novelty score diminishes. For instance, if cells are selected proportional to their novelty, the best controller of two that have novelty $2$ and $1$ will have a $2/3$ chance of being selected, while the same controller in a group of $3$ controllers with novelties $[2,1,1]$ will only have probability $1/2$ of being selected. To fight this phenomenon, we propose to restrict the novelty-proportional selection to the five most novel cells. Similarly, for exploitation steps, we sample either uniformly from the two highest fitness cells ($p=0.5$), which promotes exploitation from the current best cells, or uniformly from the two highest fitness cells from the last five updated ($p=0.5$), which promotes exploitation from newly discovered cells. We considered sampling cells with high performance progress, but this often led to the selection of sub-optimal controllers, as starting to move forward from a controller going backward is one of the best way to make progress.

\paragraph{Hyperparameters} We use fully connected neural network controllers with two hidden layers of size $256$, $tanh$ non-linearities, Xavier initialization \cite{xavier}, an Adam optimizer \cite{adam} with learning rate $0.01$ and a l2-coefficient of $0.005$ for regularization. We run \es for $n\_optim\_gens=10$ consecutive generations with a population size of $n=10^4$ and a noise parameter $\sigma=0.02$. After each step, we evaluate the average fitness $F(\theta_g)$, average behavioral characterization $BC(\theta_g)$ and associated novelty score $N(\theta_g)$ of $\theta_g$ over $n\_eval=30$ episodes. Novelty is computed as the average distance between the controller's $BC$ and its $k=10$ nearest neighbors in the archive of all past $BC$s, regardless of whether they were added to the $BM$. The code and environment are provided at \url{https://github.com/uber-research/Map-Elites-Evolutionary}.

\subsection{Damage Recovery}
\begin{figure}[!hbt]
  \centering
  \includegraphics[width=0.85\columnwidth]{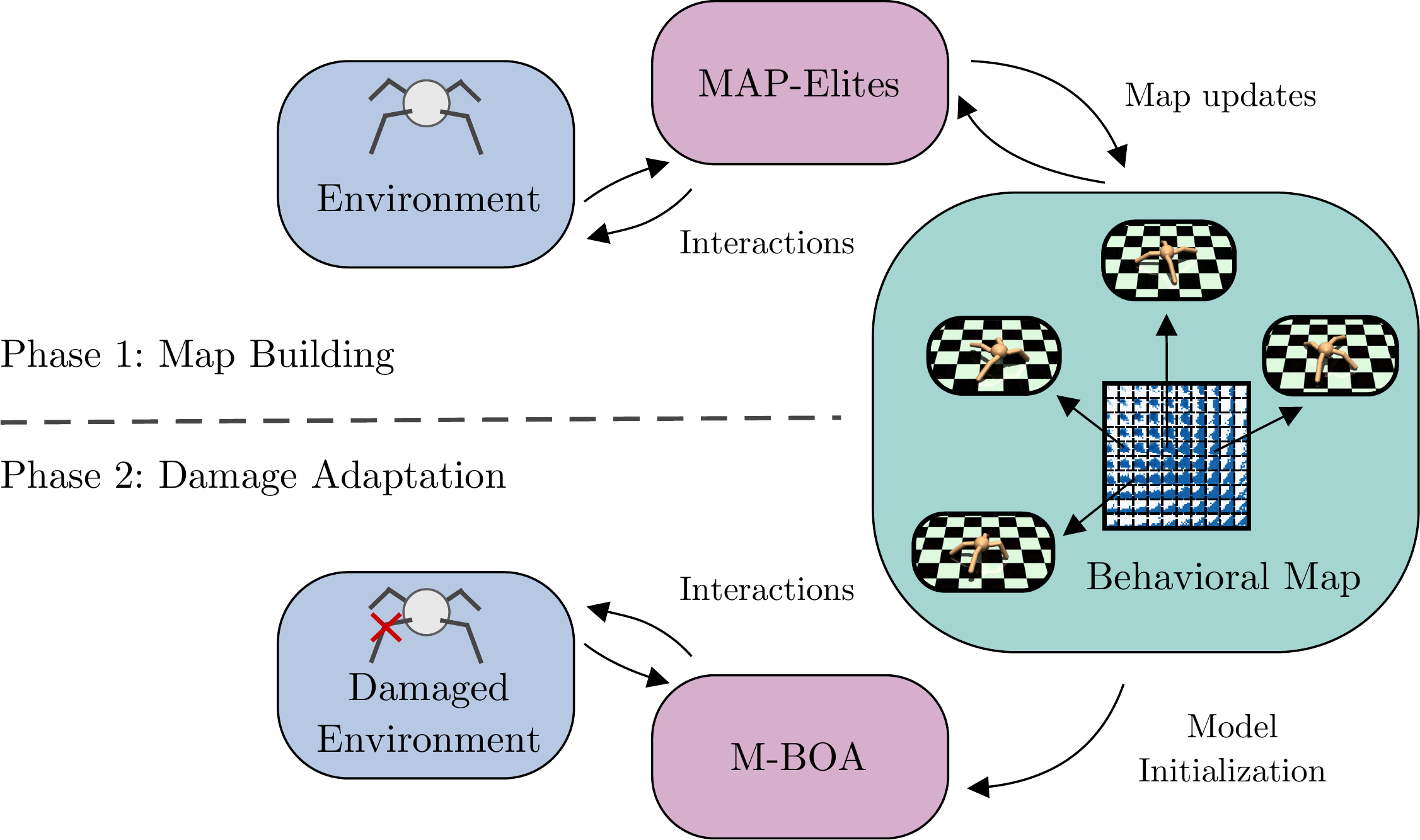}
  \caption{Building repertoires for damage adaptation. Phase~1: \me builds a repertoire of diverse and high-performing behaviors. Phase~2: \mboa builds a model of the objective function under the perturbed conditions to find a controller robust to the damage.}
  \label{fig:behav_exp}
\end{figure}

The Intelligent Trial and Error (IT\&E) algorithm integrates the evolution of a behavioral map using \me and a subsequent procedure enabling agents to recover from damage by searching the $BM$ for efficient recovery controllers \cite{cully2015robots}. Here, we reuse this setting to compare the traditional implementation of \me based on \ga and our proposed \mees variants, see Fig.~\ref{fig:behav_exp}. Once the behavioral map has been filled by \mega or \mees, we can use it for damage adaptation (e.g. loss of control for one or several joints). As adaptation procedure, we use the map-based Bayesian optimization algorithm (\mboa), part of the original IT\&E \cite{cully2015robots}. Bayesian optimization is especially suitable to find the maximum of an unknown objective function for which it is costly to obtain samples. We define the objective function as $\mathcal{F} : BC \to F$. \mboa initializes a model of $\mathcal{F}$ using the behavioral map filled by \me and updates the model using Gaussian Process (GP) regression. Data acquisition boils down to the evaluation of a controller $\theta$ in the perturbed environment, providing a new pair $(BC(\theta), F(\theta))$ to update the model $\mathcal{F}$. We follow the implementation of \citet{cully2015robots} and use a Mat\'ern kernel function with parameters $\nu~=~5/2$ and $\rho~=~0.03$ as well as hyperparameters $\kappa=0.3$ and $\sigma_{noise}^2=0.01$ as Upper Confidence Bound exploration parameter and noise prior respectively (see code).



\subsection{Baselines and Controls}
In this paper, we test \mees in two applications. The first is the construction of behavioral repertoires for \textit{damage adaptation} and the second is exploration in deceptive reward environments (\textit{deep exploration}). The two applications have different baselines and control treatments, which are described in the following sections.

\subsubsection{MAP-Elites with Genetic Algorithm}
We use the traditional implementation of \me based on \ga (\mega) as a baseline in the \textit{damage adaptation} experiments in order to highlight the lift from using \es instead of \ga to power \me in high-dimensional control tasks. To ensure a fair comparison, it is important to keep the number of episodes constant between \mega and \mees. Here, \mees requires $10,030$ episodes for each new controller ($10^4$ evaluations for the offsprings and $30$ evaluations of the final controller to get a robust performance estimate). As such, we enable \mega to add up to $334$ new controllers per generation, each being evaluated $30$ times.

We do not use \mega as a baseline in the \textit{deep exploration} experiments because the task builds on the Humanoid-v1 environment where \citet{deepga} demonstrated that \textsc{GA}s were not competitive with \es, even with $150$x more environment interactions.

\subsubsection{Novelty Search and Evolution Strategies}
We compare \mees to \nses and its variants \cite{nses} for the \textit{deep exploration} experiments. Since these baselines do not build behavioral repertoires, they are not used for the \textit{damage adaptation} application. \nses replaces the performance objective of \es by the same novelty objective as \meesexplore. Instead of a behavioral map, \nses and its variants keep a population of $5$ parent controllers optimized in parallel via \es. \nses never uses fitness for optimization and thus is a pure exploration algorithm. We also compare against two variants of \nses that include a fitness objective: (1) \nsres optimizes the average of the fitness and novelty objectives, while (2) \nsraes implements an adaptive mixture where the mixing weight is updated during learning. For \nsraes, the mixing weight leans towards exploration (novelty objective) when the best fitness stagnates and shifts back towards exploitation (fitness objective) when new behaviors are discovered that lead to higher performance.

The weight adaptation strategy of \nsraes was modified in this paper. In the original work, the weight starts at $1$ (pure exploitation), decreases by $0.05$ every $50$ generations if fitness does not improve and increases by $0.05$ at every improvement. In practice, we found this cadence too slow to adapt properly, as the mixing weight can only go to $0$ after $1000$ generations in the best case. Thus, we increase the update frequency to every $20$ generations and remove the bias towards exploitation at the start by setting the mixing weight to $0.5$ initially.


\section{Experiments}
Our two experiments are presented in the next sections. Recall that controllers are implemented by fully-connected neural networks with $2$ hidden layers of size $256$ (about $10^5$ parameters). This results in search spaces that have orders of magnitude more dimensions than those used in traditional \qd studies \cite{ns,mapelite,nslc,cully_qd}.  For each treatment, we report the mean and standard deviations across $5$ seeds, except for \mega, for which we present $3$ seeds. \mega is about $3$ times slower to run ($>3$ days) and does not manage to learn recovery controllers in the \textit{damage adaptation} task.

\subsection{Building Repertoires for Damage Adaptation}
\label{sec:adaptation}
We compare the quality of the $BM$ generated by the different versions of \me for damage recovery. In the first phase we build a behavioral map with each of the \me algorithms and in the second phase we use the behavioral map created in phase 1 to help the agent recover from damage to its joints via \mboa (Fig.~\ref{fig:behav_exp}). The damage applied to joints, indexed by $\mathcal{J}_{i}$, include:
 
 \begin{itemize}
     \item One joint cannot be controlled $(\mathcal{J}_{i\in[0..7]})$.
     \item One full leg cannot be controlled $(\mathcal{J}_{0,1}, \mathcal{J}_{2,3}, \mathcal{J}_{4,5}, \mathcal{J}_{6,7})$.
 \end{itemize}

\subsubsection{Domain}
We evolve a repertoire of walking gaits in the Mujoco Ant-v2 environment~\cite{brockman2016openai}. The $BC$ is a 4-D vector, where each dimension represents the proportion of episode steps where each leg is in contact with the floor ($[0, 1]^4$). This behavioral space is discretized into $10$ bins along each dimension, for a total of $10^4$ cells. The fitness is a mixture of the current $x$ position and a cost on the torque for each joint. This incentivizes the agent to move as far along the x-axis as possible, without exerting too much energy.

\begin{figure}[!h]
  \centering
  \hspace{-0.5cm}\subfigure[\label{fig:adaptation_perf}]{\includegraphics[width=0.65\columnwidth]{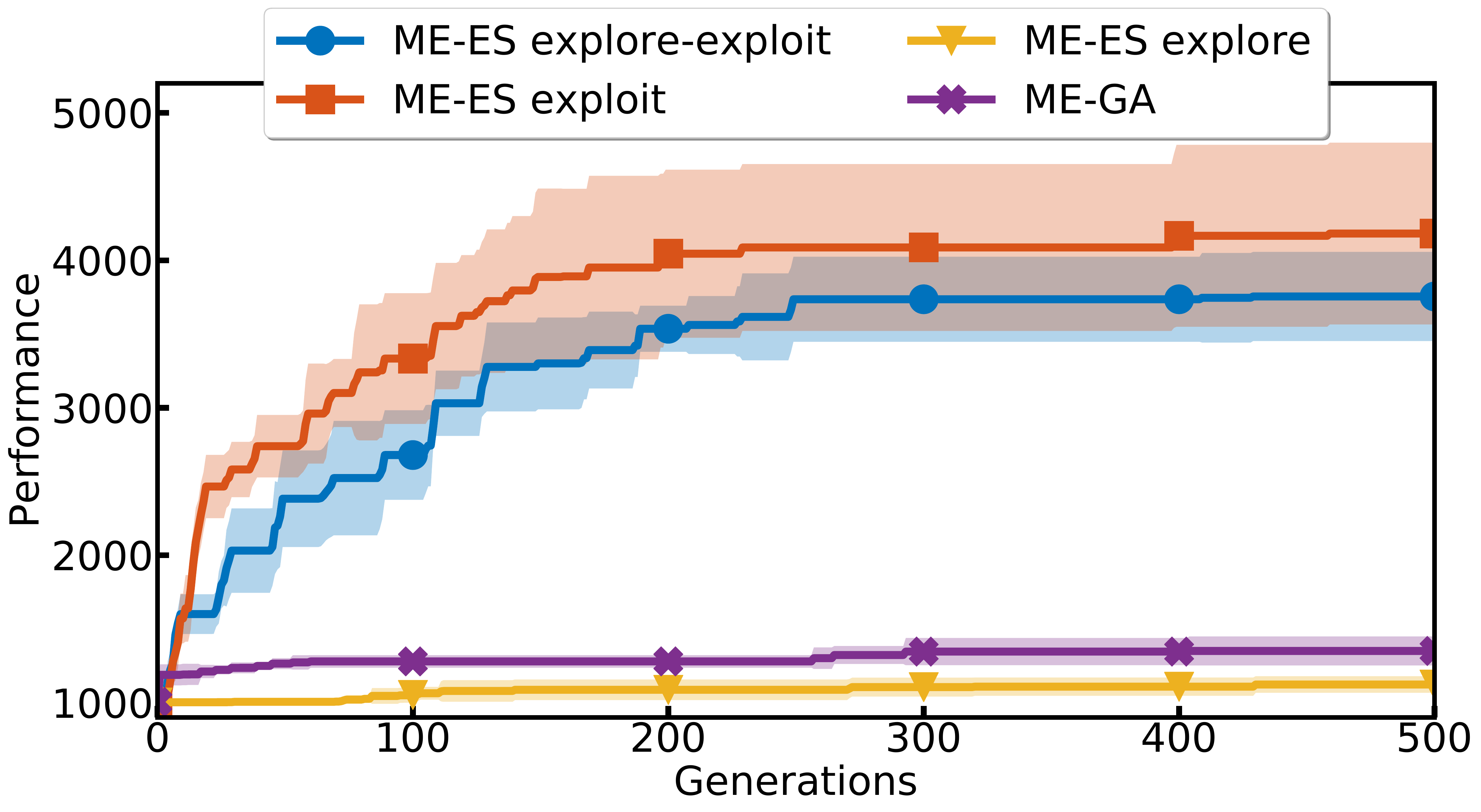}} \\
  \hspace{-0.35cm}\subfigure[\label{fig:adaptation_coverage}]{\includegraphics[width=0.65\columnwidth]{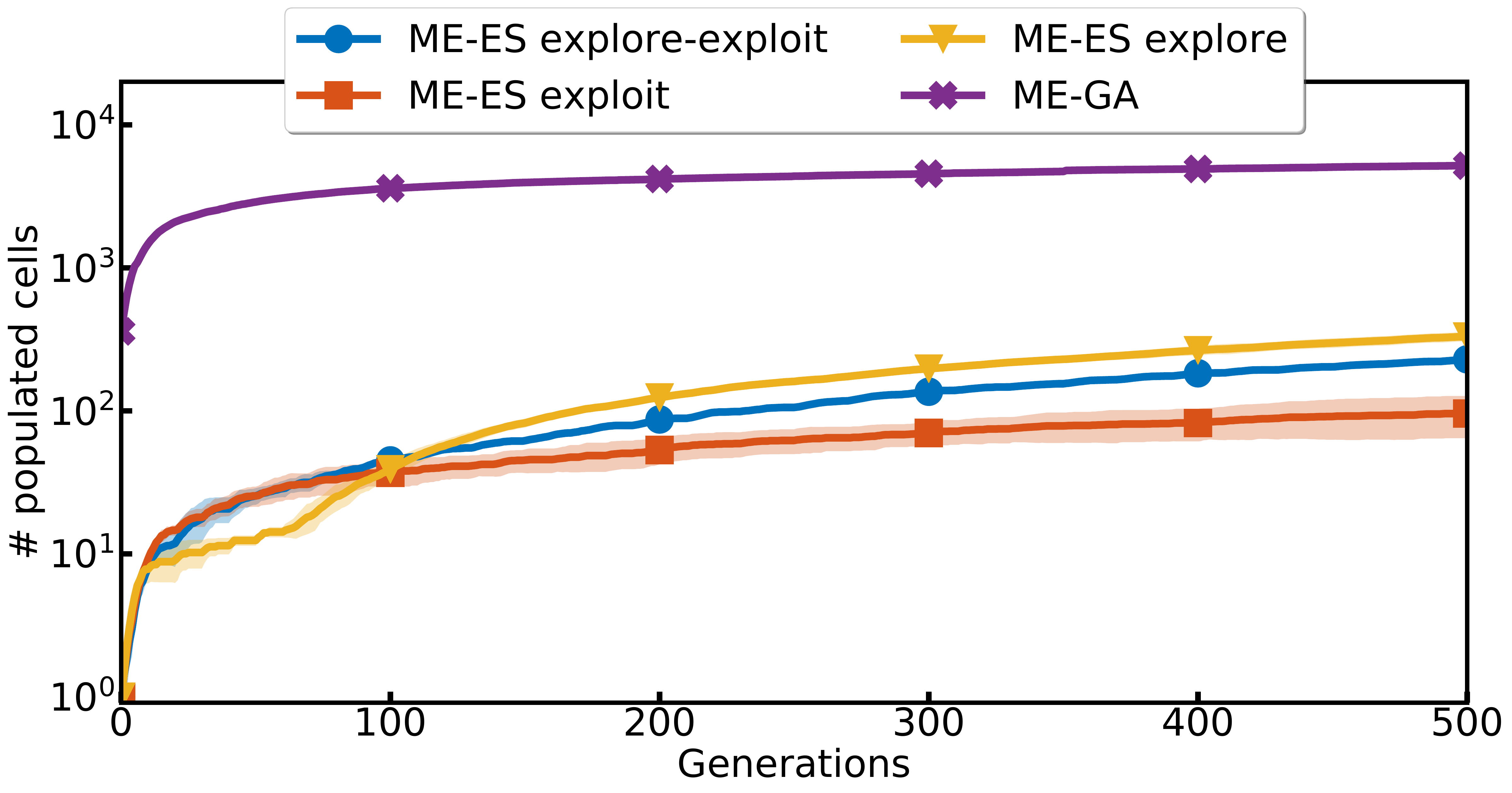}} 
  \caption{\textit{Ant} behavioral map. a: Map performance (best performance in map). b: Map coverage (\# populated cells), log scale on $y$-axis.}
  \label{fig:adaptation}
\end{figure} 

\subsubsection{Results - Map Building}
Fig.~\ref{fig:adaptation_perf} shows the evolution of the map's best performance and \ref{fig:adaptation_coverage} shows the map's coverage. As \mega adds up to $334$ times more controllers to the map per generation, its coverage is orders of magnitude higher than \mees variants. However, the best performance found in the map is quite low for \mega. \meesexplore also shows poor performances across its map. In contrast, \meesexploit and \meesexex show high performance, on par with Twin-Delayed Actor Critic ($\approx 4200$) \cite{fujimoto2018addressing} but below Soft-Actor Critic ($\approx 6000$) \cite{haarnoja2018soft}, two state-of-the-art DRL algorithms for this task.

 \begin{figure*}[!h]
  \centering
  \subfigure[\label{fig:adapt_recov}]{\includegraphics[width=0.49\textwidth]{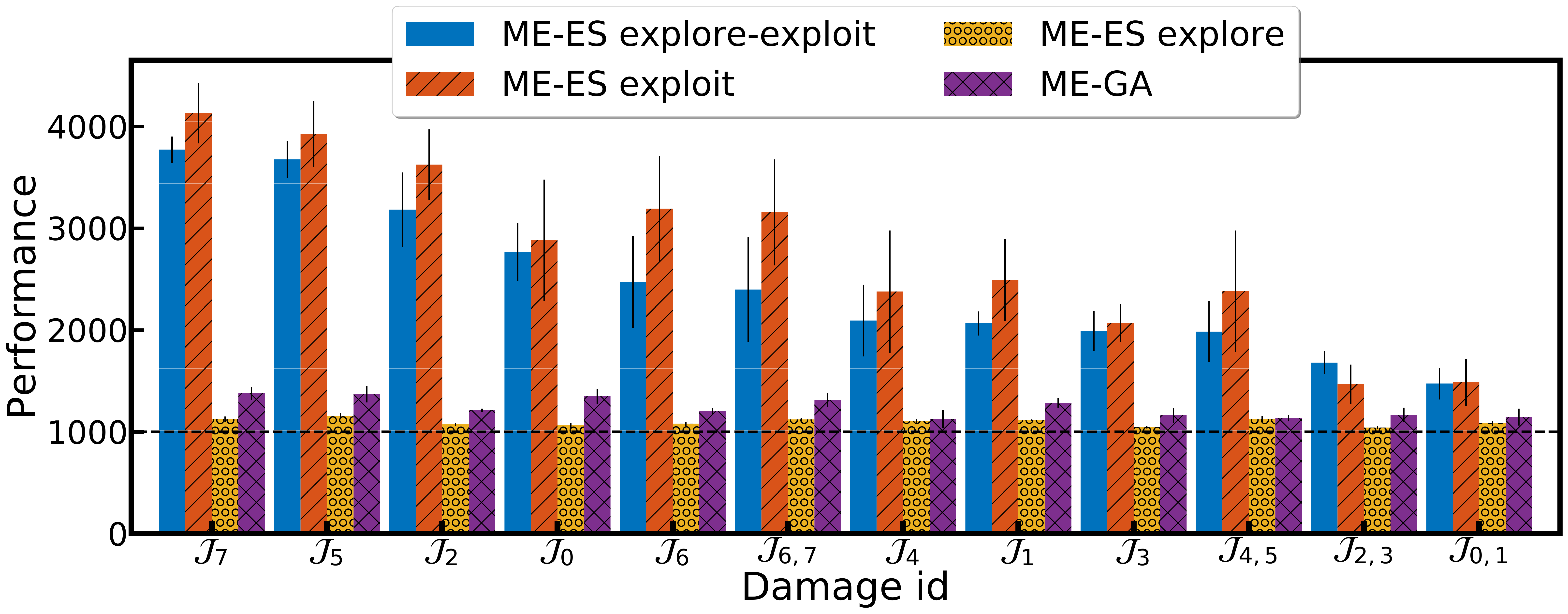}} 
  \subfigure[\label{fig:adapt_diff}]{\includegraphics[width=0.49\textwidth]{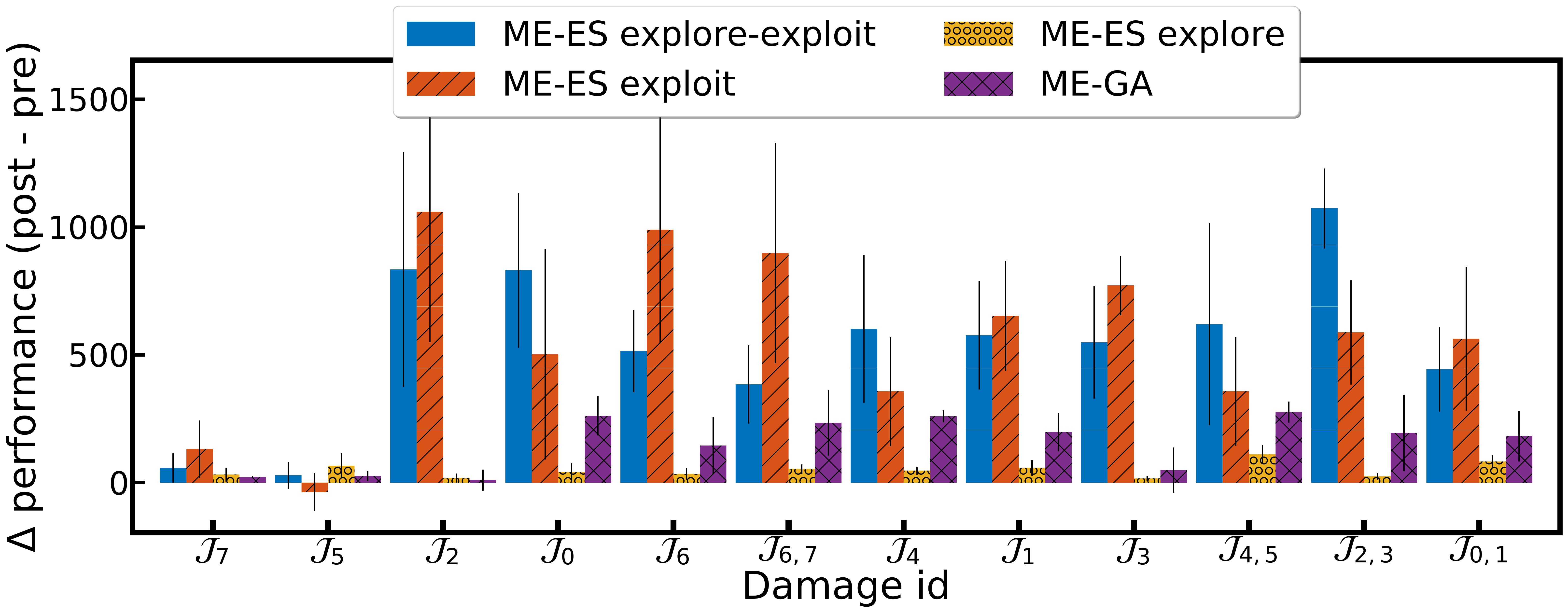}}
  \caption{Damage adaptation. a: Performance of the best recovery controller. The joint damage is represented as $\mathcal{J}_{i}$ where $i$ is the identifier of the broken joint or joints ($i \in [1..8]$). Dashed line represent the score of an agent that does not move forward but does not fall. b: Difference in performance, post- and pre-behavioral adaptation.}
  \label{fig:adapt_1leg}
\end{figure*}

\subsubsection{Results - Damage Adaptation}
\mboa is run on the behavioral maps from each experiment. Recovery controllers extracted from \mega and \meesexplore maps do not achieve high scores (Fig.~\ref{fig:adapt_recov}). The scores are around $1000$, which correspond to a motionless Ant (not moving forward, but not falling either). In contrast, \meesexex and \meesexploit are able to recover from joint damage and the ant is able to move in all damage scenarios (Fig.~\ref{fig:adapt_recov}). In Fig.~\ref{fig:adapt_diff}, we highlight the difference in performance before and after adaptation. Pre-adaptation performance is computed by evaluating the highest performing pre-damage controller in each of the post-damage environments. The post-adaption performance is computed by evaluating the recovery controller in each of the damaged joint environments. In most cases, \meesexex and \meesexploit are able to significantly improve upon the pre-damage controller (Fig.~\ref{fig:adapt_diff}). In both figures, averages and standard errors are computed over different runs of \me + \mboa (different seeds). The performance of a given recovery controller is averaged over $30$ episodes.

\subsubsection{Interpretation}
\mega does not scale to the high-di\-men\-sio\-nal Ant-v2 task, probably because it relies on random parameter perturbations (i.e. a \ga) for optimization, which does not scale to high-dimensional controllers \cite{deepga}. \meesexplore performs poorly, as it is not interested in performance, but only in exploring the behavioral space, most of which is not useful. The exploitation ability of \meesexplore only relies on its directed exploration component to find higher-performing solutions by chance. Indeed, neither \mega nor \meesexplore leverages directed exploitation. \meesexploit only focuses on exploitation, but still performs undirected exploration by retaining novel solutions that fill new cells. While \meesexex targets performance only half of the time, it still finds a $BM$ that is good for damage recovery. Its exploration component enables better coverage of the behavioral space while its exploitation component ensures high performing controllers are in the map. Note that a good $BC$ space coverage is a poor indicator of adaptation performance, whereas having a high-performing, somewhat populated map is a good indicator. Directed exploitation seems to be required to achieve good performance in the \textit{Ant-v2} task. Undirected exploration --as performed by \meesexploit-- is sufficient to build a behavioral map useful for damage recovery, as its recovery performance is similar to that of variants using directed exploration and exploitation (\meesexex).

\FloatBarrier

\subsection{Deep Exploration}
\label{sec:explo}
\subsubsection{Domains}
\begin{figure}[!hbt]
  \centering
  \subfigure[\label{fig:ant}]{\includegraphics[width=0.52\columnwidth]{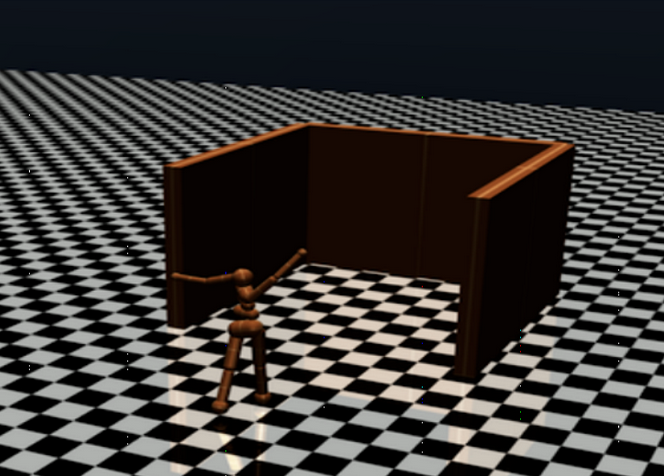}}
  \subfigure[\label{fig:humanoid}]{\includegraphics[width=0.45\columnwidth]{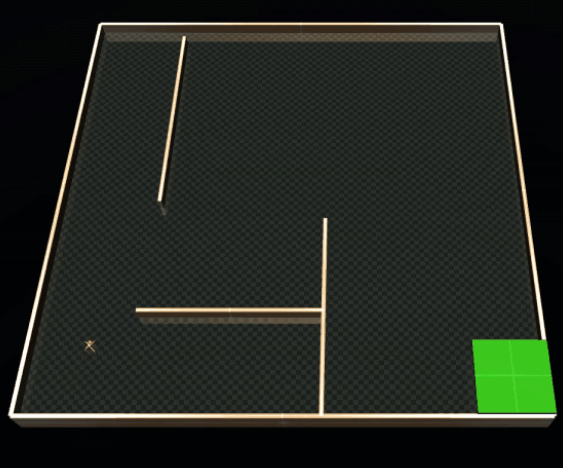}}
  \caption{\textbf{Domains for the deep exploration study.} a: \textit{Deceptive Humanoid} domain. b: \textit{Ant Maze} domain. Here the goal is the green area and the trap is in the cul-de-sac to the right of the agent's starting position.}
    \label{fig:domains}
\end{figure}

We use two modified domains from OpenAI Gym based on \textit{Humanoid-v2} and \textit{Ant-v2} (Fig.~\ref{fig:domains}). In \textit{Deceptive Humanoid}, the humanoid robot faces a U-shaped wall (like in \cite{nses}), while in \textit{Ant Maze} the Ant is placed in a maze similar to \cite{frans2017meta}. Both environments possess a strongly deceptive reward, whose gradient leads the agent directly into a trap. For both environments, we use the final $(x,y)$ position of the agent as the BC. In \textit{Deceptive Humanoid}, the fitness is defined as the cumulative reward, a mixture of velocity along the $x$-axis and costs for joint torques. See \citet{brockman2016openai} for more details. In \textit{Ant Maze}, the fitness is the Euclidean distance to the goal (green area). Controllers are run for $3000$ timesteps in \textit{Ant Maze} and up to $1000$ timesteps in \textit{Deceptive Humanoid}, less if the agent falls over. 

\subsubsection{Results - \textit{Deceptive Humanoid}}
\begin{figure}[hb!]
  \centering
  \hspace{-0.5cm}\subfigure[\label{fig:humanoid_perf}]{\includegraphics[width=0.75\columnwidth]{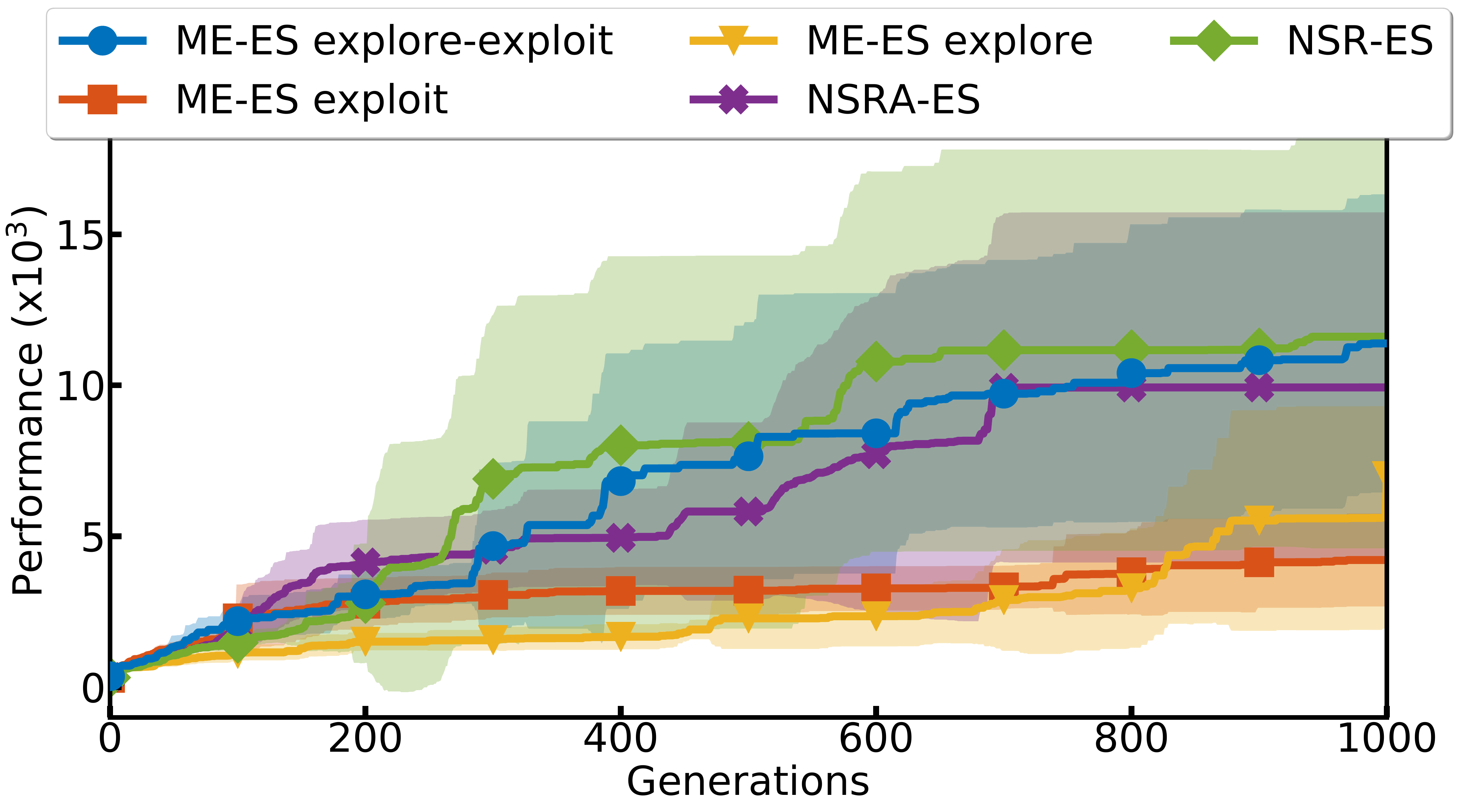}} \\
  \hspace{-0.5cm}\subfigure[\label{fig:humanoid_coverage}]{\includegraphics[width=0.76\columnwidth]{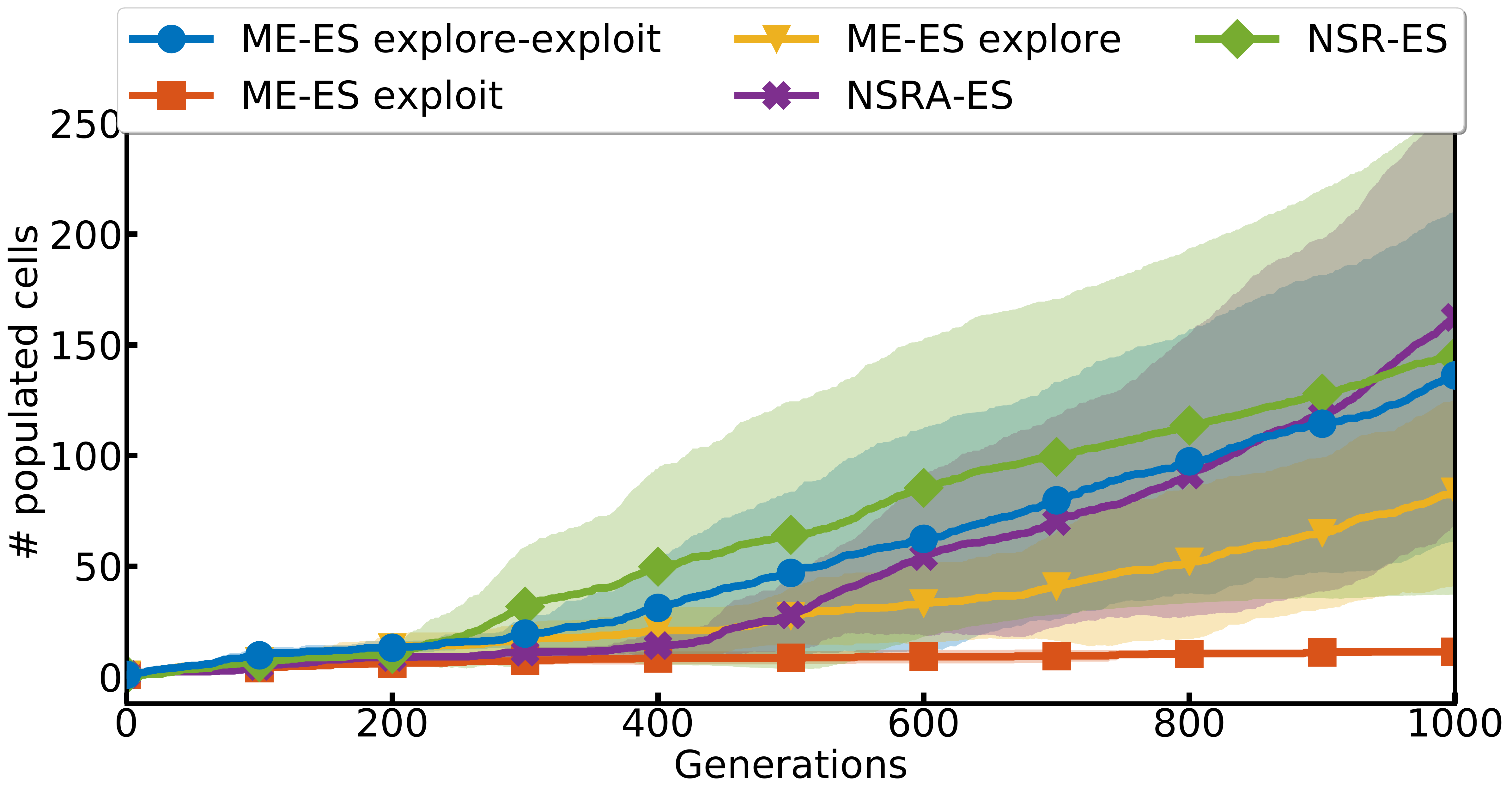}} 
  \caption{\textit{Deceptive Humanoid} behavioral map. a: Map performance (best  in map). Being stuck in the trap corresponds to around $3000$. b: Map coverage (\# populated cells).}
  \label{fig:res_humanoid}
\end{figure}

Undirected exploration (\meesexploit) is insufficient to get out of the trap and directed exploration alone (\meesexplore) falls short as well (Fig.~\ref{fig:humanoid_perf}). That said, like \nses, \meesexplore eventually explores around the wall, as indicated by its greater than $3000$ performance. Algorithms implementing directed exploration and directed exploitation manage to go around the wall and achieve high performance (Fig.~\ref{fig:humanoid_perf}), regardless of whether they use decoupled (\meesexex) or coupled (\nsres, \nsraes) exploration-exploitation. Surprisingly, \meesexplore displays poor map coverage despite its focus on exploration (Fig.~\ref{fig:humanoid_coverage}).

\subsubsection{Results - \textit{Ant Maze}}

\begin{figure}[!hbt]
  \centering
  \includegraphics[width=0.65\columnwidth]{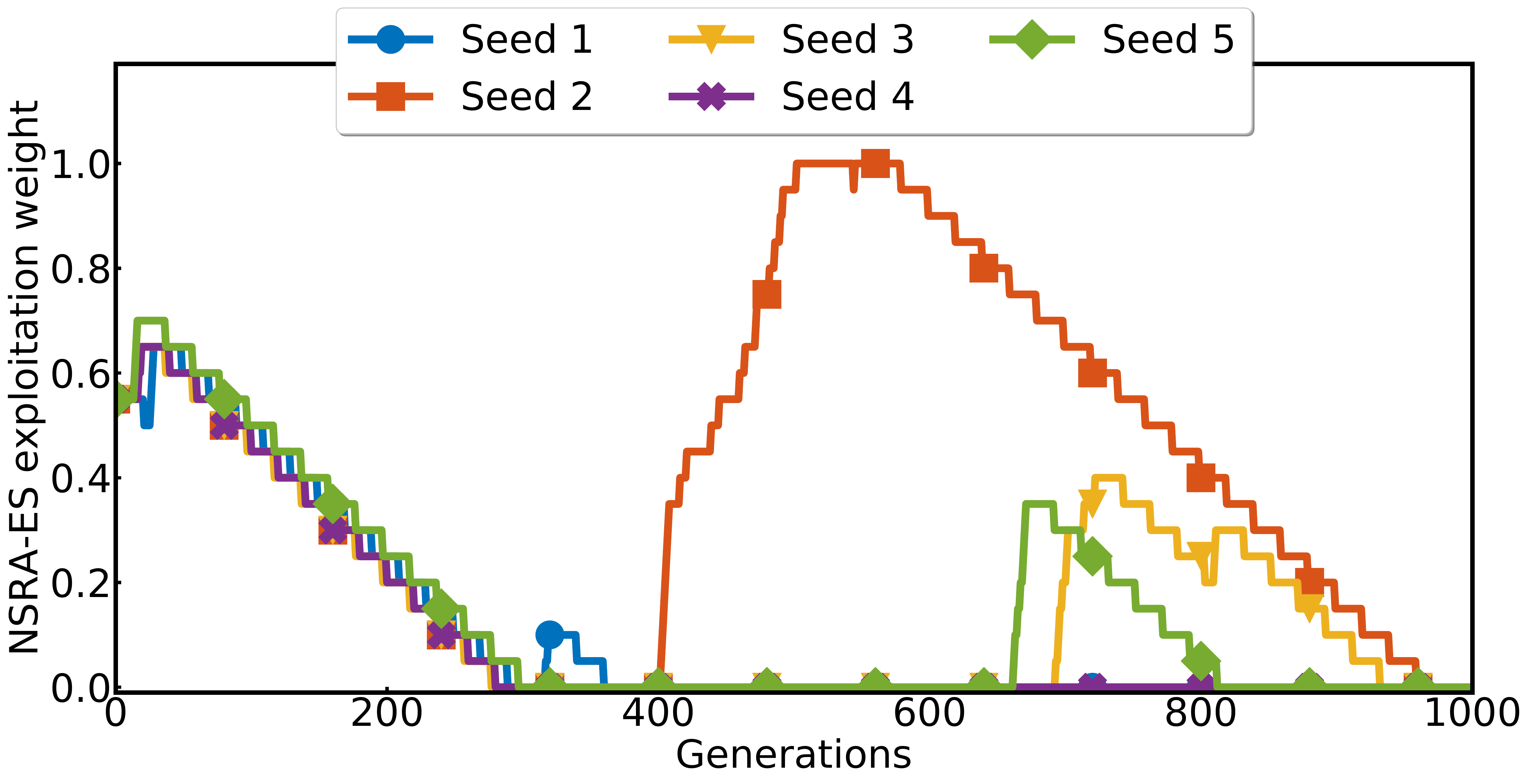}
  \caption{\textit{\nsraes weight adaptation}. Evolution of \nsraes weight tuning the ratio between the performance (exploitation) and the novelty (exploration) objectives for $5$ runs. Higher values correspond to higher levels of exploitation.}
  \label{fig:nsraes}
\end{figure}

In the \textit{Ant Maze}, both \meesexploit and \nsres get stuck in the deceptive trap, as indicated by a score of $-27$. All other methods (\meesexplore, \nses, \meesexex, and \nsraes) are able to avoid the trap and obtain a score closer to $0$. Examining the exploitation ratio of \nsraes, we observe that all runs quickly move towards performing mostly exploration. That said, some runs have a ratio that falls to pure exploration and stays there, whereas in others the algorithm manages to find a high-performing controller, which triggers an increase in preference towards performance (Fig.~\ref{fig:nsraes})

\begin{figure}[!h]
  \centering
   \hspace{-0.5cm}\subfigure[\label{fig:antmaze_perf}]{\includegraphics[width=0.75\columnwidth]{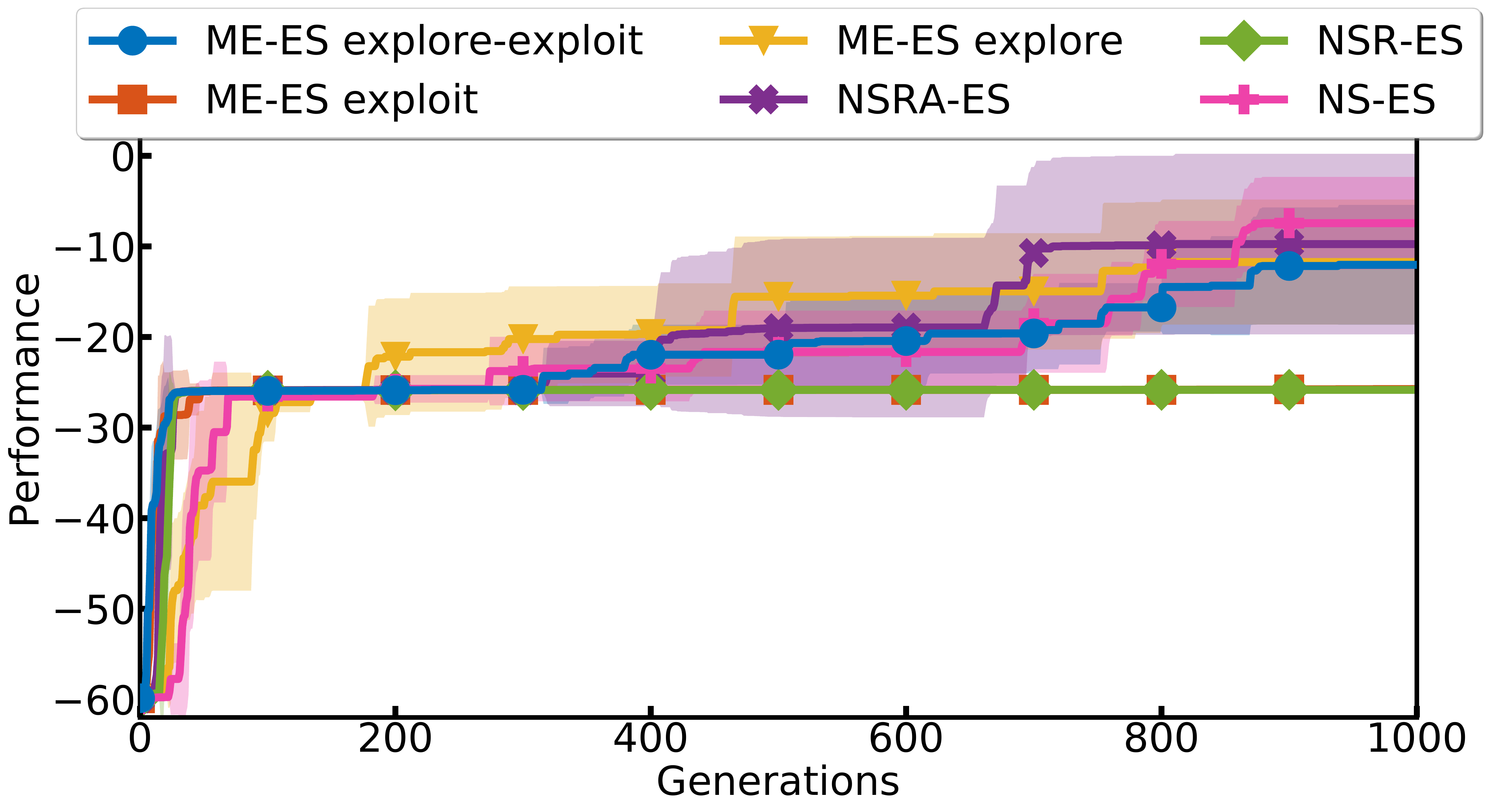}} \\
  \hspace{-0.35cm}\subfigure[\label{fig:antmaze_coverage}]{\includegraphics[width=0.74\columnwidth]{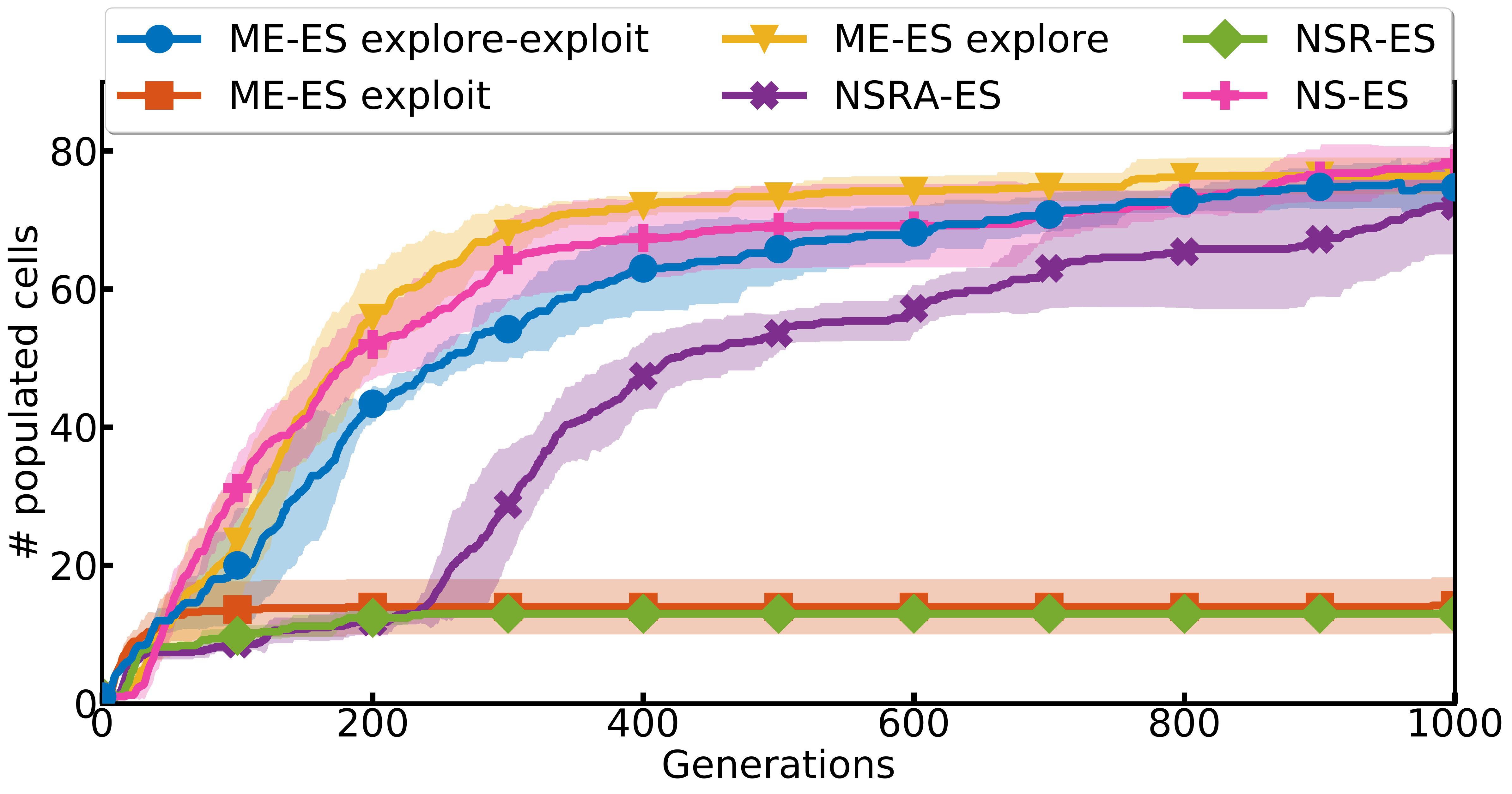}} 
  \caption{\textit{Ant Maze} behavioral map. a: Map performance (negative distance to the goal area in map). The trap corresponds to $-27$. b: Map coverage (\# populated cells).}
  \label{fig:res_ant_maze}
\end{figure}

\subsubsection{Interpretation}
\paragraph{Deceptive Humanoid.} In this environment, simply following the fitness objective can only lead the agent into the U-shaped trap. Because the environment is open-ended, pure exploration does not guarantee the discovery of a high performing behavior either, as it is possible to endlessly explore in directions misaligned with performance, a phenomenon previously encountered with \nses in \citet{nses}. This explains the poor performance of both \meesexplore and \meesexploit. However, combining exploration and exploitation provides the best of both worlds, and we observe that all algorithms that do so navigate around the wall (\nsres, \nsraes, \meesexex). As there is no need to follow gradients orthogonal to those of the performance objective to succeed, coupling exploration and exploitation (e.g. \nsres) is sufficient to achieve high fitness. The poor map coverage of  \meesexplore may be a result of it only optimizing for performance indirectly via \me map updates. As a result, it is never encouraged to learn to walk efficiently, and thus unable to fill large portions of the map. In \citet{nses} however, \nses manages to make the humanoid walk using pure directed exploration. This may be because \nses only updates a population of $5$ controllers, whereas any controller in the behavioral map of \meesexplore can be updated. A given \meesexplore controller will thus be updated less frequently than any of the $5$ \nses controllers. This dilution of the number of updates of each controller might explain why \meesexplore is less efficient at exploring than \nses in this environment. Because the environment is unbounded, different runs may explore different direction of the BC space, resulting in high standard deviations. 

\paragraph{Ant Maze.} In the Ant Maze task, pure exploitation will lead the agent to the deceptive trap (Fig.~\ref{fig:ant}). However, the environment is enclosed by walls so the agent cannot explore endlessly. Thus, performing pure exploration will eventually lead to the goal. This explains the success of directed exploration algorithms (\nses, \meesexplore) and the poor performance of directed exploitation algorithms (\meesexploit). This environment requires a more extensive exploration procedure to achieve the goal than \textit{Deceptive Humanoid}. In \textit{Ant Maze}, the agent needs to avoid two traps and needs to make turns to achieve its goal, while in \textit{Humanoid Deceptive}, the agent can simply walk with an angle of $45$ degrees to achieve high scores. As such, an even mix of the performance and novelty objectives (\nsres) fails, as we try to average two contradicting forces. \nsraes has a better chance of getting out of the trap by adding more and more exploration to the mix until the agent escapes. In doing so, however, the mixing weight can go down to $w=0$ and only consider the exploration objective, thus losing sight of the goal (Fig.~\ref{fig:nsraes}). \meesexex is able to escape the trap because of its decoupled objectives; exploration steps will lead it to explore the environment until it reaches a point where exploitation steps allow it to reach the goal.


\section{Discussion and Conclusion}
In this paper we present \mees, a new \qd algorithm scaling the powerful \me algorithm to deep neural networks by leveraging \es. We present three variants of this algorithm: (1) \meesexploit, where \es targets a performance objective, (2) \meesexplore, where \es targets a novelty objective, and (3) \meesexex, where \es targets performance and novelty objectives in an alternating fashion. Because the \me behavioral map implicitly implements undirected exploration (update rule 1) and undirected exploitation (update rule 2), these variants implement directed exploitation with undirected exploration, directed exploration with undirected exploitation, and decoupled directed exploration and exploitation, respectively.

In a first application, we use \mees to curate an archive of diverse and high-performing controllers that is subsequently used for damage adaptation. We show that \mees manages to recover from joint damage in a high-dimensional control task while \mega does not. In a second application, we use \mees to solve strongly deceptive exploration tasks. Here, we show that \meesexex performs well in both open and closed domains, on par with the state-of-the-art exploration algorithm \nsraes. As in most DRL control tasks, the controllers optimized in this work contain about $10^5$ parameters \cite{haarnoja2018soft,es}. Previous work showed \es could scale to even larger controllers (e.g. $\approx 10^6$ in Atari games) \cite{es}, which suggests \mees could as well.

Others have conducted work similar to ours. \citet{frans2017meta} tackles an environment similar to Ant Maze with hierarchical RL. Their method pre-trains independent low-level controllers to crawl in various directions before learning a high-level control to pick directions to find the goal. On top of requiring some domain knowledge to train low-level controllers, hierarchical RL has trouble composing sub-controllers, a problem that is solved by resetting the agent to a default position between sub-sequences. In this work, however, we learn to solve the deep exploration problem by directly controlling joints. 

Concurrently to our work, \citet{mecma} proposed a similar algorithm called \cmame, in which \me is combined with \cmaes, another algorithm from the \es family \cite{hansen2016cma}. In addition to a performance objective, \cmame considers two others. The \textit{improvement} objective selects for controllers that discovered new cells or improved over already populated cells. The \textit{random direction} objective rewards controllers for minimizing the distance between their $BC$ and a behavioral target vector randomly sampled from the behavioral space. Although the last objective does encourage exploration, it is not explicitly directed towards novelty as in \meesexplore and is not decoupled from performance. Additionally, although \cmaes is generally recognized as a very powerful algorithm, in practice it is limited to comparatively low-dimensional parameter spaces due to the algorithmic complexity induced by its covariance matrix updates $(O(n^2))$.

Because \mees only affects the optimization procedure of \me, any application leveraging behavioral maps could benefit from its improved optimization efficiency. In addition to the demonstrated applications for exploration (Section~\ref{sec:explo}) or damage adaptation (Section~\ref{sec:adaptation}), previous works have used \me for maze navigation \cite{pugh2015confronting} or to explore the behavioral space of soft robotic arms and the design space of soft robots \cite{mapelite}. It can also co-evolve a repertoire of diverse images matching each of the ImageNet categories \cite{nguyen2015innovation} or generate diverse collections of $3D$ objects \cite{lehman2016creative}. 

\mees could also leverage advances proposed for either \me or \es. \citet{vassiliades2017using}, for example, propose to use a centroidal Voronoi tessellation to scale \me to high-dimensional behavioral spaces, while \citet{chatzilygeroudis2018reset} improve on the behavioral adaptation procedure to enable adaptation without the need to reset the environment (or for humans to intervene in the case of real robots). Finally, \citet{evoes} recently proposed to replace the performance objective of \es by an \textit{evolvability} objective, which aims to maximize the effect of parameter perturbations on behavior. Evolvability can make a good exploration objective because perturbing \textit{evolvable} controllers is likely to increase the speed at which a behavioral map can be explored.

More generally, \mees could leverage a set of various objectives in parallel. As the bottleneck lies in the computation of the $10^4$ offspring, various objectives can be computed and optimized for simultaneously (novelty, performance, combinations of them as in \nsres, evolvability etc.). Each resulting controller can be tested $30$ times and considered a candidate for the behavioral map updates, at a negligible increase in cost of $N_{objectives} \times 30\ll10^4$.

In the \textit{deep exploration} experiment, \mees could also be extended to perform hierarchical evolution. In such an algorithm, solutions are \textit{chains} of controllers, where each link is a controller run for a given amount of time. After sampling a cell and retrieving the associated chain, the algorithm can either mutate the last controller of the chain or add a new controller to it. This would implement the \textit{go-explore} strategy proposed in \citet{ecoffet2019go}, where the agent first returns to a behavioral cell before exploring further.

The addition of \mees and its variants to the \qd family opens new avenues, as the exploration and diversity properties of \qd algorithms can now be scaled to high-dimensional problems. 

\clearpage
\newpage
\textbf{Implementation.} A python implementation or \mees is available at \url{https://github.com/uber-research/Map-Elites-Evolutionary}.
\bibliography{biblio} 
\bibliographystyle{ACM-Reference-Format}

\end{document}